\documentclass{article}

\usepackage{microtype}
\usepackage{natbib}
\usepackage{graphicx}
\usepackage{subcaption}
\usepackage{multirow}
\usepackage{amsmath}
\usepackage{amssymb}
\usepackage{booktabs} % for professional tables
\usepackage{diagbox}

\usepackage{hyperref}

\usepackage[accepted]{icml2021}

\icmltitlerunning{Accelerate CNNs from Three Dimensions: A Comprehensive Pruning Framework}
\begin{document}
	
	\twocolumn[
	\icmltitle{Accelerate CNNs from Three Dimensions: A Comprehensive\\ Pruning Framework}
	
	
	
	\icmlsetsymbol{equal}{*}
	
	\begin{icmlauthorlist}
		\icmlauthor{Wenxiao Wang}{zju,tencent}
		\icmlauthor{Minghao Chen}{zju}
		\icmlauthor{Shuai Zhao}{zju}
		\icmlauthor{Long Chen}{cucnn,ailab}
		\icmlauthor{Jinming Hu}{zju}
		\icmlauthor{Haifeng Liu}{zju}
		\icmlauthor{Deng Cai}{zju}
		\icmlauthor{Xiaofei He}{zju}
		\icmlauthor{Wei Liu}{tencent}
	\end{icmlauthorlist}
	\icmlaffiliation{zju}{State Key Lab of CAD\&CG, Zhejiang University, China}
	\icmlaffiliation{ailab}{Tencent, China. This work was done when Long Chen was at Tencent}
	\icmlaffiliation{tencent}{Tencent Data Platform, China}
	\icmlaffiliation{cucnn}{Columbia University, US}
	
	\icmlcorrespondingauthor{Deng Cai}{dengcai@gmail.com}
	
	%\icmlkeywords{Machine Learning, ICML}
	
	\vskip 0.2in
	]
	
	
	\printAffiliationsAndNotice{}  % leave blank if no need to mention equal contribution
	
\begin{abstract}
		Most neural network pruning methods, such as filter-level and layer-level prunings, prune the network model along one dimension
        (\textit{depth}, \textit{width}, or \textit{resolution}) solely to meet a computational budget.
		However, such a pruning policy often leads to excessive reduction of that dimension, thus inducing a huge accuracy loss.
		To alleviate this issue, we argue that pruning should be conducted along three dimensions comprehensively.
		For this purpose, our pruning framework formulates pruning as an optimization problem.
		Specifically, it first casts the relationships between a certain model's accuracy and \textit{depth/width/resolution} into a polynomial regression
        and then maximizes the polynomial to acquire the optimal values for the three dimensions.
        Finally, the model is pruned along the three optimal dimensions accordingly.
		In this framework, since collecting too much data for training the regression is very time-costly, we propose two approaches to lower the cost:
        1) specializing the polynomial to ensure an accurate regression even with less training data;
        2) employing iterative pruning and fine-tuning to collect the data faster.
		Extensive experiments show that our proposed algorithm surpasses state-of-the-art pruning algorithms and even neural architecture search-based algorithms.
\end{abstract}

\section{Introduction}
	
  \begin{figure} \flushleft
    \includegraphics[scale=0.91]{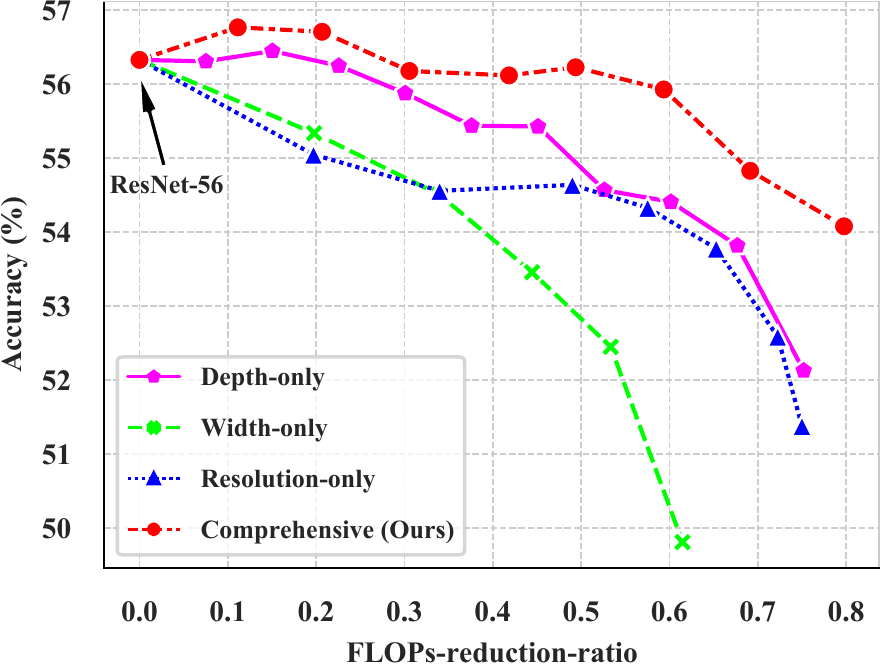}
	\caption{Accuracies of a base neural network model with different pruning policies on TinyImageNet. The base model is ResNet-56 (FLOPs-reduction-ratio = 0). $\mathcal{X}$-only means that the model is pruned only along $\mathcal{X}$ dimension, and ``comprehensive'' means that three dimensions are pruned comprehensively. A larger FLOPs-reduction-ratio implies a higher acceleration ratio. }
		\label{fig:intro}
		\vspace{-0.32cm}
	\end{figure}

To deploy pre-trained Convolutional Neural Networks (CNNs)~\cite{DBLP:journals/corr/SimonyanZ14a,DBLP:conf/cvpr/HeZRS16,DBLP:conf/cvpr/HuangLMW17,DBLP:conf/icml/TanL19} on resource-constrained mobile devices, plenty of methods~\cite{DBLP:conf/nips/BaC14,hinton2015distilling,DBLP:conf/iccv/LiuLSHYZ17,he2019filter,DBLP:conf/iclr/FrankleC19} have been proposed for model acceleration.
Among them, neural network pruning, which prunes redundant components (e.g., filters) of CNNs to cater for a computational budget,
is one of the most popular and is the focus of this paper.
	
Currently, the dominant pruning methods fall into three categories: (1) \textbf{layer-level pruning}~\cite{DBLP:journals/corr/abs-1912-10178,DBLP:conf/cvpr/LinJYZCYHD19}, which prunes redundant layers and reduces model's depth, (2) \textbf{filter-level pruning}~\cite{DBLP:conf/iccv/LiuLSHYZ17,DBLP:conf/iclr/0022KDSG17,DBLP:conf/iclr/MolchanovTKAK17,he2019filter,wang2019cop,DBLP:journals/pami/LuoZZXWL19,DBLP:conf/icml/KangH20,DBLP:conf/icml/YeGNZKL20}, which prunes redundant filters and reduces model's width, and (3) \textbf{image-level pruning}\footnote{Though images are actually resized for acceleration, we will use term ``pruning images'' for simplicity in what  follows.}~\cite{DBLP:journals/corr/HowardZCKWWAA17,DBLP:conf/icml/TanL19,han2020model}, which resizes images and reduces model's input resolution.
These three kinds of pruning methods respectively focus on one single dimension (i.e., depth, width, or image resolution) that impacts on a model's computational cost.
	
Naturally, we raise an important question overlooked by much previous work: given a pre-trained neural network model, which dimension --- depth, width, or resolution --- should we prune to minimize the model's accuracy loss?
In practice, users empirically choose a redundant dimension, which, however, often leads to a sub-optimal pruned model because of an inappropriate dimension choice. Even worse, excessive pruning of whichever dimension will cause an unacceptable loss, as shown in Figure~\ref{fig:intro}. Instead, comprehensively pruning these three dimensions yields a much lower loss than solely pruning whichever dimension, demonstrated by Figure~\ref{fig:intro}, therefore enabling model acceleration with much better quality.
	
In this paper, we propose a framework that prunes three dimensions comprehensively. Instead of solely pruning one dimension to reduce the computational cost, our framework first decides how much of each dimension should be pruned. To this end, we formulate model acceleration as an optimization problem. Precisely, given a pre-trained neural network model and a target computational cost, assuming that the pruned model's depth, width, and resolution are $d \times 100\%$, $w\times 100\%$, and $r \times 100\%$ of the original model, respectively, we seek the optimal $(d, w, r)$ that maximizes the model's accuracy -- $a$:
	\begin{equation}
	\label{equ:intro}
	\begin{aligned}
	\mathop{\max}_{d, w, r}\ a := \mathcal{F}(d,w,r), \;\; \mathrm{s.t.}\;\; \mathcal{C}(d,w,r) = \tau,
	\end{aligned}
	\end{equation}
where $\mathcal{F}(d, w, r)$ is a Model Accuracy Predictor (MAP). $\mathcal{C}(d,w,r)$ and $\tau$ represent the model's computational cost and its constraint, respectively. \cite{DBLP:conf/icml/TanL19} has designed a reasonable expression for $\mathcal{C}(d,w,r)$. However, designing a MAP manually is unachievable as its form can be arbitrarily complicated or even varies with the architecture (e.g., the MAPs for ResNet and MobileNet may be in different forms). Hence, we propose approximating the MAP via a polynomial regression, because polynomials can approximate arbitrary continuous functions according to Taylor's theorem. Specifically, we can formulate the MAP as a polynomial and collect a sufficient set of $(d, w, r, a)$ as training data to estimate its parameters. Then, problem~\eqref{equ:intro} can be solved with Lagrange's multiplier theorem, and the model is eventually pruned in terms of the optimized $(d, w, r)$.
	
The main challenge that this framework encounters is that the polynomial regression requires tremendous training data (i.e., $\{(d, w, r, a)\}$), while the collection of the data is very costly because fetching each item of data, i.e., a $(d, w, r, a)$, means training a new neural network model from scratch.
To reduce both the collection time and model training cost, we improve the framework in two aspects: 1) A specialized polynomial is proposed whose weight tensor is replaced with its low-rank substitute. The low-rank weight tensor prevents the polynomial from overfitting and ensures an accurate regression even with limited training data. Further, as a bonus, the updated MAP owns a more concise form. 2) Given a pre-trained model, we prune and fine-tune it iteratively to acquire a series of new models and their corresponding $\{(d, w, r, a)\}$, which is much faster than training such new models from scratch.
	
Extensive experiments are conducted to show the superiority of our proposed pruning algorithm over the state-of-the-art pruning algorithms.
Further, we compare against some algorithms that balance the size of three dimensions (depth, width, and resolution) from a Neural Architecture Search (NAS) perspective.
The comparative results also show our advantages over them.
	
It is worth highlighting that the contributions of this work are three-fold:
%	\vspace{-0.15cm}
	\begin{itemize}
	\item We propose to prune a model along three dimensions comprehensively and determine the optimal values for these dimensions by solving a polynomial regression and subsequently an optimization problem.
%		\vspace{-0.15cm}
		\item To complete the regression process with an acceptable cost, we apply two approaches: 1) specializing a MAP adapting to the scenario of limited training data; 2) using iterative pruning and fine-tuning to collect data faster.
%		\vspace{-0.15cm}
		\item We do extensive experiments to validate that our proposed algorithm outperforms state-of-the-art pruning and even NAS-based model acceleration algorithms.
%		\vspace{-0.25cm}
	\end{itemize}

\section{Background and Related Work}
	
\paragraph{Neural Network Pruning:} In the early stage, neural network pruning is done at the weight-level~\cite{DBLP:journals/corr/HanMD15,DBLP:conf/iclr/FrankleC19,DBLP:conf/nips/Sehwag0MJ20,DBLP:conf/eccv/YeDLGQYC20,DBLP:conf/iclr/FrankleC19,DBLP:conf/iclr/LeeAT19}. However, it needs specific libraries for sparse matrix calculation (e.g., cuSPARSE) to accelerate the inference, while these libraries' support on mobile devices is restricted. Nowadays, the most dominant pruning methods are at the filter-level, layer-level, or image-level, directly reducing the computational cost for all devices. Filter-level pruning~\cite{DBLP:conf/iccv/LiuLSHYZ17,DBLP:conf/iclr/0022KDSG17,DBLP:conf/iclr/MolchanovTKAK17,DBLP:conf/ijcai/HeKDFY18,he2019filter,wang2019cop,DBLP:conf/icml/KangH20,DBLP:conf/icml/YeGNZKL20,DBLP:conf/eccv/LiGZGT20,DBLP:conf/aaai/WangZXZSZH20} compresses models by removing unimportant filters in CNNs, layer-level pruning~\cite{DBLP:journals/corr/abs-1912-10178} does that by pruning redundant layers, and image-level pruning~\cite{DBLP:journals/corr/HowardZCKWWAA17} saves computation by using small input images. They all receive great success in pruning CNNs. However, focusing on
pruning one dimension solely also restricts their potentials.
	
%\vspace{-0.3cm}

\begin{figure*}  \centering
	\includegraphics[scale=0.488]{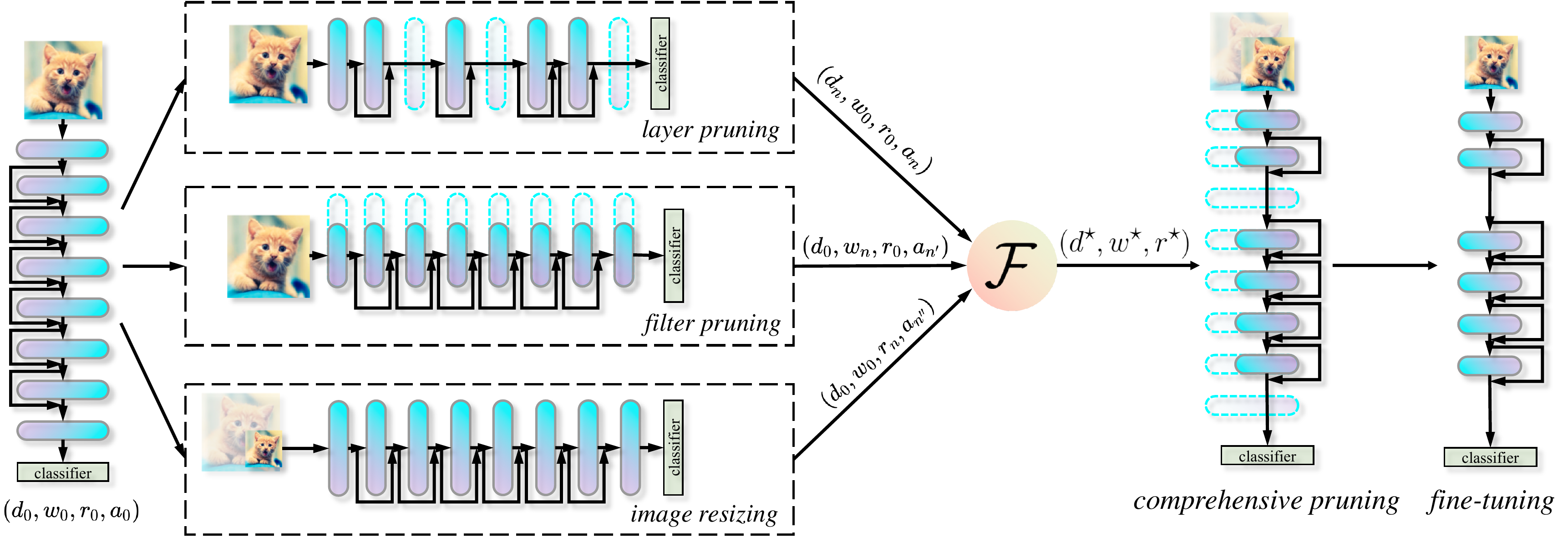}
	\caption{The pipeline of the proposed pruning framework. It first prunes a pre-trained model from three dimensions independently, yielding a set of $(d_n, w_n, r_n, a_n)$ that is taken as training data. Then, the training data is used to fit our specialized MAP ($\mathcal{F}$) via a polynomial regression. The optimal $(d^\star, w^\star, r^\star)$ is then acquired by maximizing $\mathcal{F}$ subject to a computational cost constraint. Finally, the model will be pruned comprehensively in terms of $(d^\star, w^\star, r^\star)$.}
	\label{fig:alg}
\end{figure*}
	
\paragraph{Multi-Dimension Pruning:} To the best of our knowledge, there are two methods~\cite{DBLP:conf/nips/WenWWCL16,DBLP:conf/cvpr/LinJYZCYHD19} which prune models at both the filter- and layer-levels. Both of them train models with extra regularization terms and induce sparsity into the models. Then the filters or layers with much sparsity will be pruned with a slight loss incurred. However, the same method cannot be used for balancing image size because images do not contain trainable parameters, and there is no way to induce sparsity into the images. In contrast, our proposed framework can balance three dimensions comprehensively, yielding
better model acceleration results than the above mentioned methods.
	
\vspace{-0.1cm}
	
\paragraph{Pruning vs. NAS:} Pruning and NAS~\cite{DBLP:conf/icml/PhamGZLD18,DBLP:conf/cvpr/GaoBJMJL20,DBLP:conf/cvpr/HeYSZ20,DBLP:journals/corr/abs-2006-09134,DBLP:conf/iclr/LiuSY19,DBLP:conf/icml/TanL19,han2020model,DBLP:conf/iccv/HowardPALSCWCTC19,DBLP:conf/nips/HuangCBFCCLNLWC19,DBLP:conf/iclr/ZophL17} share the same goal, that is, maximizing a certain model's accuracy given a computational budget. However, their settings are very different: pruning shrinks the model from a pre-trained one, utilizing both pre-trained model's architecture and weights, while NAS searches the model(s) from scratch. Therefore, though several algorithms~\cite{DBLP:conf/icml/TanL19,han2020model} also attempt to balance the three dimensions (i.e., depth, width, and resolution) of the model from a NAS perspective, they cannot be applied for pruning directly.

\section{Proposed Framework}
	
\subsection{Preliminaries}
	
For a model $M$, we define $\mathcal{D}(M)$, $\mathcal{W}(M, l)$, and $\mathcal{R}(M)$ as its depth, width, and input resolution. Specifically, $\mathcal{D}(M)$ represents the number of blocks\footnote{E.g., Conv-BN-ReLu blocks and residual blocks.} that $M$ contains; $\mathcal{W}(M, l)$ denotes the number of filters of a certain layer $l$ in the model $M$; $\mathcal{R}(M)$ is the side length of $M$'s input image. Given a pre-trained model $M_0$, we also define $d_n$, $w_n$, and $r_n$ of a pruned model $M_n$ as:
\begin{equation}
	\label{equ:def}
d_n=\frac{\mathcal{D}(M_n)}{\mathcal{D}(M_0)},\ w_n=\frac{\mathcal{W}(M_n, l)}{\mathcal{W}(M_0, l)},\ r_n = \frac{\mathcal{R}(M_n)}{\mathcal{R}(M_0)}.
\end{equation}
For filter pruning, following previous work~\cite{DBLP:journals/pami/LuoZZXWL19,he2019filter,DBLP:conf/cvpr/LinJWZZ0020}, we prune all layers with the same ratio, so $w_n$ of a model has no concern with the choice of layer $l$. Further, for a pruning task, it is easy to know: $d_n, w_n, r_n \in (0, 1]$ and $d_0 = w_0 = r_0 = 1$.
	
The pipeline of our proposed pruning framework is introduced in Figure~\ref{fig:alg}. Unlike previous work that prunes one dimension solely, we first look for a pruning policy (i.e., how much of each dimension should be pruned) which aims to maximize the model's accuracy in Section~\ref{sec:maao}. Then, we depict the process of pruning and fine-tuning a target model in terms of pruning policy in Section~\ref{sec:pruning}.

\subsection{Model Acceleration as Optimization} \label{sec:maao}
	
\subsubsection{Formulation} \label{sec:formulation}

Given a CNN architecture, the model's depth, width, and image resolution are three key aspects that affect both the model's accuracy and its computational cost. 
Thus, model acceleration can be formulated as the following problem:
\begin{equation}
	\label{equ:method1}
	\begin{aligned}
	d^\star, w^\star, r^\star &= \mathop{\arg\max}\limits_{d, w, r} \mathcal{F}(d, w, r;\Theta) \\
	&\quad \ \ \mathrm{s.t.}\; \mathcal{C}(d, w, r) = T \times\mathcal{C}(d_0, w_0, r_0), % 0\le d \le d_0; 0 \le w \le w, 0 \le r
	\end{aligned}
\end{equation}
where $\mathcal{F}(d, w, r;\Theta)$ is a \textbf{Model Accuracy Predictor (MAP)}, which predicts the model's accuracy given $(d, w, r)$. $\Theta$ contains the parameters of the MAP. $\mathcal{C}(d, w, r)$ represents the computational cost (e.g., FLOPs) of a model. $T \in (0, 1)$ implies that the pruned model's computational cost is $T$ proportion of the original model. Problem \eqref{equ:method1} can be solved using Lagrange's multiplier theorem once $\mathcal{F}(d, w, r)$ and $\mathcal{C}(d, w, r)$ are known. Following \cite{DBLP:conf/icml/TanL19} in which a model's computational cost is proportional to $d, w^2$, and $r^2$, we re-define $\mathcal{C}(d, w, r)$ as:
\begin{equation}
	\label{equ:method1_11}
	\begin{aligned}
	\mathcal{C}(d, w, r) = dw^2r^2.
	\end{aligned}
\end{equation}
However, designing a MAP manually is unachievable as its form can be arbitrarily complicated, and different architectures may own different forms. An intuitive idea is resorting to a polynomial regression because any continuous function can be approximated with polynomials according to Taylor's theorem. Specifically, we can train $N$ models with different $(d, w, r)$, attaining their accuracy $a$, and fit the MAP with a polynomial by using $\{(d_n, w_n, r_n, a_n)\}_{n=1}^N$ as training data.
However, the regression process requires hundreds of data items $(d_n, w_n, r_n, a_n)$ for training an accurate regression, whereas fetching each item of that data needs us to  train a new model from scratch, which is very resource-inefficient and time-consuming. To overcome this obstacle, on the one hand, we specialize a MAP that ensures an accurate regression even with less training data in Section~\ref{sec:simplified-MAP}. On the other hand, we expedite acquiring each data item $(d_n, w_n, r_n, a_n)$ by employing iterative pruning and fine-tuning in Section~\ref{sec:fast-data-collection}.

\subsubsection{Specialized MAP} \label{sec:simplified-MAP}

The polynomial-shaped MAP can be represented as:
\begin{equation}
	\label{equ:tmp1}
	\begin{aligned}
	\mathcal{F}(d, w, r; \Theta) = \sum_{i, j, k=0}^{\mathcal{K}}\theta_{ijk}d^iw^jr^k,
	\end{aligned}
\end{equation}
where $\Theta \in \mathbb{R}^{(\mathcal{K}+1)\times (\mathcal{K}+1)\times (\mathcal{K}+1)}$ is a tensor, and all $\theta_{ijk}$ are its elements\footnote{For the convenience of expression, we assume the same highest degree $\mathcal{K}$ for $d$, $w$, and $r$, though conclusions in this section still hold when they have different $\mathcal{K}$.}. Without any constraint on $\Theta$, the polynomial can be highly flexible and expressive. However, high flexibility also makes it easy to overfit~\cite{bishop:2006:PRML}, especially when the training data (i.e., $\{(d, w, r, a)\}$) is scarce. To avoid overfitting and ensure an accurate regression with limited training data, a relatively simple MAP with less flexibility and expressiveness is needed. We achieve this by restricting the rank\footnote{The definition of tensor rank is the same as \cite{bourbaki2003elements}.} of $\Theta$ during the regression process, i.e., $\Theta$ in the MAP is replaced by its low-rank substitute. Formally, for $\Theta$ of rank $\widetilde{\mathcal{R}}$, its $\mathcal{R}$-rank substitute ($\mathcal{R} < \widetilde{\mathcal{R}}$) and elements are defined as~\cite{DBLP:journals/siamrev/KoldaB09}:
\begin{equation}
	\label{equ:tmp2}
	\begin{aligned}
	\Theta & \approx \sum_{q=1}^\mathcal{R} \vec{s_q} \otimes \vec{u_q} \otimes \vec{v_q}, \ \ \ \theta_{ijk} \approx \sum_{q=1}^\mathcal{R}s_{qi}u_{qj}v_{qk}, \\
	\end{aligned}
\end{equation}
in which $\otimes$ represents outer product, and $\vec{s_q}, \vec{u_q}, \vec{v_q} \in \mathbb{R}^{\mathcal{K}+1}$ represent $(\mathcal{K}+1)$-dimensional vectors, e.g., $\vec{s_q} = [s_{q0}, s_{q1}, \cdots, s_{q\mathcal{K}}]^\top$. Then, replacing $\theta_{ijk}$ in Eq.~\eqref{equ:tmp1} yields:
\begin{equation}
	\label{equ:tmp3}
	\begin{aligned}
	\mathcal{F}(d, w, r; \Theta) &\approx \sum_{i, j, k=0}^{\mathcal{K}}\sum_{q=1}^\mathcal{R}s_{qi}u_{qj}v_{qk} d^iw^jr^k \\
	&= \sum_{q=1}^\mathcal{R}\sum_{i, j, k=0}^{\mathcal{K}}(s_{qi}d^{i})(u_{qj}w^j)(v_{qk}r^k) \\
	&=\sum_{q=1}^\mathcal{R} \sum_{i=0}^{\mathcal{K}}s_{qi}d^i \sum_{j=0}^{\mathcal{K}}u_{qj}w^j \sum_{k=0}^{\mathcal{K}}v_{qk} r^k \\
	&=\sum_{q=1}^\mathcal{R} \mathcal{H}_{}(d; \vec{s_q})\mathcal{H}_{}(w; \vec{u_q})\mathcal{H}_{}(r; \vec{v_q}),
	\end{aligned}
\end{equation}
in which $\mathcal{H}$ represents univariate polynomial. In practice, we take Eq.~\eqref{equ:tmp3} as our MAP and control its flexibility by adjusting $\mathcal{R}$. A smaller $\mathcal{R}$ indicates a simpler MAP. Empirically, we find that $\mathcal{R}=1$ is enough for achieving an accurate regression in most cases, which provides our MAP with a highly succinct form. We also verify through experiments that $\mathcal{R}=1$ makes sense because it accords with the prior of the MAP (Section~\ref{sec:ablation}).

	\begin{algorithm}[tb]
		\caption{\quad Iterative Pruning and Fine-tuning}
		\label{alg:fast-data-collection}
		\begin{algorithmic}
			\STATE {\bfseries input} pre-trained $M_0$, rounds $rds$, pruning setting $T$
			\STATE {\bfseries initialize} $train\_data = \{(d_0, w_0, r_0, a_0)\}$
			%		\STATE $d = \frac{d_0 - d_{min}}{rs}, w = \frac{w_0 - w_{min}}{rs}, r = \frac{r_0 - r_{min}}{rs}$.
%			\For{$X$ in $\{d, w, r\}$}
			\FUNCTION{PruneAlong($dimension$, $x_0$, $x_{min}$)}
%				\STATE {\bfseries use global} $M_0$, $train\_data$, $rds$
				\FOR{$n=1$ {\bfseries to} $rds$}
				\STATE $x_n = x_{n-1} - \frac{x_0 - x_{min}}{rds}$
				\STATE {\bfseries pruning} $M_{n-1}$ {\bfseries along} $dimension$ {\bfseries to} $x_n \rightarrow M_n$
				\STATE {\bfseries fine-tuning} $M_n \rightarrow (d_n, w_n, r_n, a_n)$
				\STATE {\bfseries add} $(d_n, w_n, r_n, a_n)$ {\bfseries to} $train\_data$
				\ENDFOR
			\ENDFUNCTION
%			\STATE 
			\STATE $d_{min} = T d_0, w_{min} = \sqrt{T}w_0, r_{min}=\sqrt{T} r_0$
			\STATE PruneAlong(``depth'', $d_0$, $d_{min}$)
			\STATE PruneAlong(``width'', $w_0$, $w_{min}$)
			\STATE PruneAlong(``resolution'', $r_0$, $r_{min}$)
			\STATE {\bfseries return} $train\_data$
		\end{algorithmic}
	\end{algorithm}

\subsubsection{Fast Data Collection} \label{sec:fast-data-collection}
	
To collect data used for MAP's regression, instead of training many models with different $(d, w, r)$ from scratch, we apply iterative pruning and fine-tuning to acquire the data.
	
\paragraph{Iterative Pruning and Fine-tuning:} As shown in Algorithm~\ref{alg:fast-data-collection}, the pre-trained model $M_0$ is pruned along three dimensions independently. At each dimension, we iteratively apply pruning and fine-tuning on $M_0$ to generate many models, and the configurations $\{(d_n, w_n, r_n, a_n)\}$ of these models are collected for the MAP's regression. $d_{min}$ in Algorithm~\ref{alg:fast-data-collection} indicates that if we reduce the model's depth to $d_{min}$, the computational cost constraint $T$ can be fulfilled without pruning the model's width and input resolution. It is easy to deduce that the optimal $d^\star \ge d_{min}$. Likewise, $w_{min}$ and $r_{min}$ in Algorithm~\ref{alg:fast-data-collection} are minimal possible values for $w$ and $r$, respectively.
	
Compared with training models from scratch, our data collection strategy enjoys two advantages: 1) A pruned pre-trained model converges much faster than the one training from scratch, thus taking much less time to obtain a new model. 2) Besides the finally pruned model, iterative pruning yields several intermediate models as well as their configurations $\{(d_n, w_n, r_n, a_n)\}$, which can also be used for the MAP's regression.
	
\subsubsection{Optimizing the MAP} \label{dd}
	
With the collected data, we fit the MAP by using a regression algorithm. Then, the optimal $(d^\star, w^\star, r^\star)$ satisfies Eq.~\eqref{equ:do2} according to Lagrange's multiplier theorem, where $\lambda$ is the Lagrange multiplier.
\begin{equation}
	\label{equ:do2}
	\left\{
	\begin{aligned}
	dw^2r^2 - T \times d_0w_0^2r_0^2 &= 0 \\
	\sum\nolimits_{q=1}^{\mathcal{R}}\mathcal{H}'(d;\vec{s_q})\mathcal{H}(w;\vec{u_q})\mathcal{H}(r;\vec{v_q})+\lambda w^2r^2 &= 0 \\
	\sum\nolimits_{q=1}^{\mathcal{R}}\mathcal{H}(d;\vec{s_q})\mathcal{H}'(w;\vec{u_q})\mathcal{H}(r;\vec{v_q})+2\lambda dwr^2 &= 0 \\
	\sum\nolimits_{q=1}^{\mathcal{R}}\mathcal{H}(d;\vec{s_q})\mathcal{H}(w;\vec{u_q})\mathcal{H}'(r;\vec{v_q})+2\lambda dw^2r &= 0 \\
	\end{aligned}
	\right.
\end{equation}

\subsection{Comprehensive Pruning and Fine-tuning} \label{sec:pruning}
	
Leveraging the optimal $(d^\star, w^\star, r^\star)$, filter-level pruning and layer-level pruning are applied to prune a pre-trained model $M_0$ to the target $d^\star$ and $w^\star$, and then the model is fine-tuned with images of size $r^\star$. During the entire pruning process, layer-pruning first and filter-pruning first are both viable and yield the same pruned model. Without loss of generality, we describe the pruning process by assuming layer-pruning first, and the concrete steps are as follows:
	
\paragraph{Pruning Layers:} Following DBP~\cite{DBLP:journals/corr/abs-1912-10178}, we put a linear classifier after each layer of model $M_0$ and test its accuracy on the evaluation dataset. The accuracy of each linear classifier indicates the discrimination of its corresponding layer's features. Further, each layer's discrimination enhancement compared with its preceding layer is seen as the importance of the layer. With this importance metric, we pick out the least important $(1-d^\star/d_0) \times 100\%$ layers and remove them from $M_0$, yielding $M_{p_1}$.
	
\paragraph{Pruning Filters:} Filter-level pruning is performed over $M_{p_1}$. In particular, we use the scaling factor of BN layers as the importance metric, just like Slimming~\cite{DBLP:conf/iccv/LiuLSHYZ17}. However, different from Slimming that compares the importances of all filters globally, we only compare the importances of filters in the same layer, and the least important $(1-w^\star/w_0) \times 100\%$ filters of each layer will be pruned. Through such a modification, the pruned ratios of all layers are kept the same. Assume the model after filter-pruning to be $M_{p_2}$.
	
\paragraph{Fine-tuning with Smaller Images:} After pruning, the pruned model $M_{p_2}$ is fine-tuned with images of size $r^\star$. The images are resized by bilinear down-sampling, which is the most common down-sampling scheme for images. The model will be fine-tuned with a small learning rate till convergence, leading to the finally pruned model $M_p$.

\section{Experiments}

\subsection{Experimental Settings}

\paragraph{Datasets:} We take three popular datasets as testbeds of our algorithm: CIFAR-10~\cite{krizhevsky2009learning}, TinyImageNet~\cite{tinyimagenet}, and ImageNet~\cite{DBLP:journals/ijcv/RussakovskyDSKS15}. These three datasets differ in their image-resolutions (32$\times$32 to 224$\times$224), number of classes (10 to 1000), and scale of datasets (50K to 1000K images). For all the datasets, images are augmented by symmetric padding, random clipping, and randomly horizontal flip, all of which are common~\cite{DBLP:conf/cvpr/HeZRS16,DBLP:journals/corr/HowardZCKWWAA17,wang2019cop} augmentation methods for these datasets.

\paragraph{Architectures:} We test our algorithm on three popular network architectures: ResNet~\cite{DBLP:conf/cvpr/HeZRS16}, DenseNet~\cite{DBLP:conf/cvpr/HuangLMW17}, and EfficientNet~\cite{DBLP:conf/icml/TanL19}. Their basic blocks vary from residual blocks to densely connected blocks and NAS-searched blocks, representing three of the most popular designs for deep CNNs.

\paragraph{Evaluation Protocol:} Following the conventions of previous work~\cite{DBLP:conf/eccv/LiGZGT20,DBLP:conf/cvpr/LinJWZZ0020,DBLP:conf/icml/YeGNZKL20}, we take the accuracy, parameters-reduction-ratio ($Prr$), and FLOPs-reduction-ratio ($Frr$) as the evaluation protocol of our model acceleration algorithm. $Prr$ and $Frr$ are defined as Eq.~\eqref{equ:eval_proto}, where $M_0$ and $M_p$ represent the base model and the pruned model, respectively.
\begin{equation}
	\label{equ:eval_proto}
	Prr = 1 - \frac{\textit{Params}(M_p)}{\textit{Params}(M_0)}, Frr = 1 - \frac{\textit{FLOPs}(M_p)}{\textit{FLOPs}(M_0)}.
\end{equation}

\paragraph{Compared Algorithms:} The compared algorithms fall into three categories:
(1) Algorithms solely pruning the model along one dimension (i.e., depth, width, or resolution), including $\mathcal{R}$-only~\cite{DBLP:journals/corr/HowardZCKWWAA17}, $\mathcal{W}$-only~\cite{DBLP:conf/iccv/LiuLSHYZ17}, FPGM~\cite{he2019filter}, DBP~\cite{DBLP:journals/corr/abs-1912-10178}, PScratch~\cite{DBLP:conf/aaai/WangZXZSZH20}, DHP~\cite{DBLP:conf/eccv/LiGZGT20}, and HRank~\cite{DBLP:conf/cvpr/LinJWZZ0020};
(2) Algorithms that prune along multi-dimensions, such as GAL~\cite{DBLP:conf/cvpr/LinJYZCYHD19};
(3) NAS-based algorithms, including EfficientNet~\cite{DBLP:conf/icml/TanL19} and TinyNet~\cite{han2020model}, 
which balance the size of the three dimensions from the NAS perspective.

\begin{table*}[t]
	\caption{Pruning results on CIFAR-10 and TinyImageNet. $\mathcal{D}$, $\mathcal{W}$, and $\mathcal{R}$ indicate whether the model will be pruned along depth, width, and resolution dimension, respectively. ``Acc. Drop'' means the accuracy loss induced by pruning (smaller is better). Results with $\dagger$ are drawn from original papers, and the others are run with their published code with slight modifications. Our algorithm achieves \textbf{smaller accuracy losses than the others with similar $Prr$ and $Frr$}.}
	\begin{center}
		\scalebox{0.734}{
			\setlength{\tabcolsep}{3.5mm}{
				\begin{tabular}{c|c|l|ccc|ccccc}
					\toprule
					Dataset & Architecture& Algorithm & $\mathcal{D}$ & $ \mathcal{W} $ & $ \mathcal{R} $& Baseline & Accuracy & Acc. Drop & $Prr$ & $Frr$ \\
					\midrule
					\multirow{23}{*}{CIFAR-10} & \multirow{7}{*}{ResNet-32} & \multicolumn{1}{l|}{$\mathcal{R}$-only~\cite{DBLP:journals/corr/HowardZCKWWAA17}}& & & \multicolumn{1}{l|}{\checkmark} & 93.18\% & 90.19\% & 2.99\% & - & 0.52\\
					& & $\mathcal{W}$-only~\cite{DBLP:conf/iccv/LiuLSHYZ17} & & \checkmark & \multicolumn{1}{l|}{}& 93.18\% & 92.16\% & ~1.02\% & 0.47& 0.47\\
					& & $\mathcal{D}$-only DBP~\cite{DBLP:journals/corr/abs-1912-10178} & \checkmark & & \multicolumn{1}{c|}{}& 93.18\% & 92.65\% & ~0.53\% & 0.28& 0.48\\
					& & GAL~\cite{DBLP:conf/cvpr/LinJYZCYHD19} & \checkmark & \checkmark & \multicolumn{1}{c|}{}& 93.18\% & 91.72\% & ~1.46\% & 0.39& 0.50\\
					& & FPGM$^\dagger$~\cite{he2019filter} & \multicolumn{1}{c}{}& \checkmark & \multicolumn{1}{l|}{}& 92.63\% & 92.31\% & ~0.32\% & - & 0.42\\
					& & PScratch~\cite{DBLP:conf/aaai/WangZXZSZH20} & \multicolumn{1}{c}{}& \checkmark & \multicolumn{1}{c|}{}& 93.18\% & 92.18\% & ~1.00\% & - & 0.50\\
					& & Ours& \checkmark & \checkmark & \multicolumn{1}{l|}{\checkmark} & 93.18\% & \textbf{93.27\%} & \textbf{-0.09\%} & 0.38 & 0.49\\ \cmidrule{2-11}
					& \multirow{9}{*}{ResNet-56} & \multicolumn{1}{l|}{$\mathcal{R}$-only~\cite{DBLP:journals/corr/HowardZCKWWAA17}}& & & \multicolumn{1}{l|}{\checkmark} & 93.69\%& 92.00\% & ~1.69\% & - & 0.51\\
					& & \multicolumn{1}{l|}{$\mathcal{W}$-only~\cite{DBLP:conf/iccv/LiuLSHYZ17}}& & \checkmark & \multicolumn{1}{l|}{}& 93.69\%& 92.97\% & ~0.72\% & 0.50& 0.50\\
					& & \multicolumn{1}{l|}{$\mathcal{D}$-only DBP~\cite{DBLP:journals/corr/abs-1912-10178}} & \checkmark & & \multicolumn{1}{c|}{}& 93.69\%& 93.27\% & ~0.42\% & 0.40& 0.52\\
					& & \multicolumn{1}{l|}{GAL$^\dagger$~\cite{DBLP:conf/cvpr/LinJYZCYHD19}} & \checkmark & \checkmark & \multicolumn{1}{c|}{}& 93.26\%& 93.38\% & \textbf{-0.12\%}& 0.12& 0.38\\
					& & \multicolumn{1}{l|}{FPGM$^\dagger$~\cite{he2019filter}}& \multicolumn{1}{c}{}& \checkmark & \multicolumn{1}{l|}{}& 93.59\%& 93.26\% & ~0.33\% & - & 0.52\\
					& & \multicolumn{1}{l|}{PScratch$^\dagger$~\cite{DBLP:conf/aaai/WangZXZSZH20}} & \multicolumn{1}{c}{}& \checkmark & \multicolumn{1}{c|}{}& 93.23\%& 93.05\% & ~0.18\% & - & 0.50\\
					& & \multicolumn{1}{l|}{HRank$^\dagger$~\cite{DBLP:conf/cvpr/LinJWZZ0020}} & \multicolumn{1}{c}{}& \checkmark & \multicolumn{1}{c|}{}& 93.26\%& 93.17\% & ~0.09\% & 0.42 & 0.50\\
					& & \multicolumn{1}{l|}{DHP~\cite{DBLP:conf/eccv/LiGZGT20}} & \multicolumn{1}{c}{}& \checkmark & \multicolumn{1}{c|}{}& 93.65\%& 93.58\% & ~0.07\% & 0.42 & 0.49\\
					& & \multicolumn{1}{l|}{Ours}& \checkmark & \checkmark & \multicolumn{1}{l|}{\checkmark} & 93.69\%& \textbf{93.76\%} & \textbf{-0.07\%}& 0.40& 0.50 \\ \cmidrule{2-11}
					& \multirow{7}{*}{DenseNet-40} & \multicolumn{1}{l|}{$\mathcal{R}$-only~\cite{DBLP:journals/corr/HowardZCKWWAA17}}& & & \multicolumn{1}{l|}{\checkmark} & 94.59\%& 92.88\% & ~1.71\% & - & 0.53\\
					& & $\mathcal{W}$-only~\cite{DBLP:conf/iccv/LiuLSHYZ17}& & \checkmark & \multicolumn{1}{l|}{}& 94.59\%& 94.26\% & ~0.33\% & 0.65& 0.65\\
					& & $\mathcal{D}$-only DBP~\cite{DBLP:journals/corr/abs-1912-10178} & \checkmark & & \multicolumn{1}{c|}{}& 94.59\%& 94.02\% & ~0.57\% & 0.60& 0.46\\
					& & GAL$^\dagger$~\cite{DBLP:conf/cvpr/LinJYZCYHD19} & \checkmark & \checkmark & \multicolumn{1}{c|}{}& 94.81\%& 94.50\% & ~0.31\% & 0.57& 0.55\\
					& & HRank$^\dagger$~\cite{DBLP:conf/cvpr/LinJWZZ0020} & \multicolumn{1}{c}{}& \checkmark & \multicolumn{1}{c|}{}& 94.81\%& 93.68\% & ~1.13\% & 0.54& 0.61\\
					& & DHP$^\dagger$~\cite{DBLP:conf/eccv/LiGZGT20} & \multicolumn{1}{c}{}& \checkmark & \multicolumn{1}{c|}{}& 94.74\%& 93.94\% & ~0.80\% & 0.36& 0.62\\
					& & Ours& \checkmark & \checkmark & \multicolumn{1}{l|}{\checkmark} & 94.59\%& \textbf{94.54\%} & ~\textbf{0.05\%} & 0.66& 0.66\\
					\hline
					\multirow{18}{*}{TinyImageNet} & \multirow{6}{*}{ResNet-56} & $\mathcal{R}$-only~\cite{DBLP:journals/corr/HowardZCKWWAA17} &  &  & \checkmark & 56.55\% & 54.64\% & ~1.91\% & - & 0.49 \\
					&  & $\mathcal{W}$-only~\cite{DBLP:conf/iccv/LiuLSHYZ17} &  & \checkmark &  & 56.55\% & 52.45\% & ~4.10\% & 0.54 & 0.53 \\
					&  & $\mathcal{D}$-only DBP~\cite{DBLP:journals/corr/abs-1912-10178} & \checkmark &  &  & 56.55\% & 55.57\% & ~0.98\% & 0.25 & 0.53 \\
					&  & GAL~\cite{DBLP:conf/cvpr/LinJYZCYHD19} & \checkmark & \checkmark &  & 56.55\% & 55.87\% & ~0.68\% & 0.32 & 0.52 \\
					&  & DHP~\cite{DBLP:conf/eccv/LiGZGT20} &  & \checkmark &  & 56.55\% & 55.82\% & ~0.73\% & 0.46 & 0.55 \\
					&  & Ours & \checkmark & \checkmark & \checkmark & 56.55\% & \textbf{56.04\%} & ~\textbf{0.51\%} & 0.34 & 0.59 \\ \cmidrule{2-11}
					& \multirow{6}{*}{ResNet-101} & $\mathcal{R}$-only~\cite{DBLP:journals/corr/HowardZCKWWAA17} &  &  & \checkmark & 64.83\% & 55.48\% & ~9.35\% & - & 0.75 \\
					&  & $\mathcal{W}$-only~\cite{DBLP:conf/iccv/LiuLSHYZ17} &  & \checkmark &  & 64.83\% & 63.47\% & ~1.36\% & 0.75 & 0.75 \\
					&  & $\mathcal{D}$-only DBP~\cite{DBLP:journals/corr/abs-1912-10178} & \checkmark &  &  & 64.83\% & 61.35\% & ~3.48\% & 0.76 & 0.77 \\
					&  & GAL~\cite{DBLP:conf/cvpr/LinJYZCYHD19} & \checkmark & \checkmark &  & 64.83\% & 64.33\% & ~0.50\% & 0.45 & 0.76 \\
					&  & DHP~\cite{DBLP:conf/eccv/LiGZGT20} &  & \checkmark &  & 64.83\% & 64.82\% & ~0.01\% & 0.50 & 0.75 \\
					&  & Ours & \checkmark & \checkmark & \checkmark & 64.83\% & \textbf{65.27\%} & \textbf{-0.44\%} & 0.51 & 0.75 \\ \cmidrule{2-11}
					& \multirow{6}{*}{DenseNet-100} & $\mathcal{R}$-only~\cite{DBLP:journals/corr/HowardZCKWWAA17} &  &  & \checkmark & 61.34\% & 56.97\% & ~4.37\% & - & 0.75 \\
					&  & $\mathcal{W}$-only~\cite{DBLP:conf/iccv/LiuLSHYZ17} &  & \checkmark &  & 61.34\% & 59.56\% & ~1.78\% & 0.75 & 0.75 \\
					&  & $\mathcal{D}$-only DBP~\cite{DBLP:journals/corr/abs-1912-10178} & \checkmark &  &  & 61.34\% & 58.44\% & ~2.90\% & 0.65 & 0.78 \\
					&  & GAL~\cite{DBLP:conf/cvpr/LinJYZCYHD19} & \checkmark & \checkmark &  & 61.34\% & 59.03\% & ~2.31\% & 0.78 & 0.70 \\
					&  & DHP~\cite{DBLP:conf/eccv/LiGZGT20} &  & \checkmark &  & 61.34\% & 59.40\% & ~1.94\% & 0.73 & 0.73 \\
					&  & Ours & \checkmark & \checkmark & \checkmark & 61.34\% & \textbf{60.22\%} & ~\textbf{1.12\%} & 0.73 & 0.75 \\
					\bottomrule
			\end{tabular}}
		}
	\end{center}
	\label{tab:all}
\end{table*}
	
\vspace{-0.057cm}
\paragraph{Training Settings:} For base models trained on CIFAR-10, we set batch size to $64$ for DenseNet and $128$ for ResNet, respectively. Weight decay is set to $10^{-4}$. The models are trained for $160$ epochs with the learning rate starting from $0.1$ and divided by $10$ at epochs $80$ and $120$. These are all the most common training settings~\cite{DBLP:conf/cvpr/HeZRS16,DBLP:journals/corr/HowardZCKWWAA17,wang2019cop} for models trained on CIFAR-10. For ResNet and DenseNet trained on TinyImageNet and ImageNet, batch size is set to $256$, and weight decay is $10^{-4}$. Models are trained for $100$ epochs. The learning rate is set to $0.1$ at the beginning and is multiplied
by $0.1$ at epochs $30$, $60$, and $90$. For EfficientNet, we apply the same training policy as~\cite{han2020model}, which is also the most common for EfficientNet implemented with PyTorch~\cite{paszke2017automatic}.
	
\vspace{-0.057cm}
\paragraph{Regression and Pruning Settings:} The MAP's hyper-parameters are set to $\mathcal{R}=1$ and $\mathcal{K}=3$ in our pruning experiments. When collecting training data (i.e., $\{(d_n, w_n, r_n)\}_{n=1}^N$) for the polynomial regression, the model is pruned along each dimension for four times (i.e., $rds=4$ in Algorithm~\ref{alg:fast-data-collection}). ResNet and DenseNet trained on CIFAR-10 are fine-tuned for $40$ epochs at each round, and for $80$ epochs after comprehensive pruning. Therefore, the data collection process consumes as much time as training $3$ models (training one model from scratch costs $160$ epochs). Similarly, models trained on TinyImageNet and ImageNet are fine-tuned for $30$ epochs at each round of the iterative pruning process. Thus, it takes about the same time as training $3.6$ models for the data collection process. The finally pruned models trained on TinyImageNet and ImageNet are fine-tuned for $60$ epochs after comprehensive pruning.

\subsection{Results and Analyses} \label{exp:re_and_ana}
	
\paragraph{Results on CIFAR-10 and TinyImageNet:}
	
The experimental results on CIFAR-10 and TinyImageNet are shown in Table~\ref{tab:all}. As we can see, $\mathcal{W}$-only induces greater loss than $\mathcal{D}$-only for ResNet-32 and ResNet-56, while for ResNet-101, the situation is opposite. In other words, the importance of different dimensions lies in the original size of \textit{depth}, \textit{width}, and \textit{resolution}, and we cannot deduce it from a simple prior, which further shows the essentiality of our algorithm. We balance the size of the three dimensions dynamically and always achieve better results than pruning one or two dimensions. The most competitive opponent of our algorithm is DHP, which achieves similar accuracy and $Frr$ to our algorithm for ResNet-56 trained on both datasets. However, we show higher accuracy than DHP for DenseNet-40 on CIFAR-10 (94.54\% vs. 93.94\%), for ResNet-101 (65.27\% vs. 64.82\%) on TinyImageNet, and for DenseNet-100 (60.22\% vs. 59.40\%) on TinyImageNet with similar $Prr$ and $Frr$, which sheds light on the robustness of our algorithm across different architectures and datasets.
\vspace{-0.05cm}
	
\paragraph{Results on ImageNet:}

Experiments with ImageNet are done on ResNet-50 and DenseNet-121. From Table~\ref{tab:imagenet}, we can see that our algorithm achieves 0.45\% higher accuracy on ResNet-50 than the state-of-the-art algorithms (DHP and PScratch) with the same $Frr$. The improvement on DenseNet-121 is marginal compared with $\mathcal{W}$-only, because our algorithm also prunes DenseNet-121 mainly along width dimension, which indicates that DenseNet-121's width is \emph{large} and has much redundancy. By contrast, images do not need to be pruned. With a comprehensive consideration, our algorithm also deems that we should mainly prune filters of DenseNet-121 for acceleration. Therefore, it produces similar pruning results to filter-level pruning. However, the results do not imply that our algorithm is powerless. On the contrary, \textbf{a pruning policy with a comprehensive consideration is always better than an arbitrary one, though they may produce similar results sometimes}.

\paragraph{Comparison with NAS:}
	
Algorithms that balance the three dimensions (i.e., depth, width, and resolution) in a NAS manner are also compared, and the results are shown in Table~\ref{tab:nas}. GPU-days is the most common metric to evaluate the search cost of NAS algorithms, which indicates natural days they spend if running with only one GPU. Both EfficientNet~\cite{DBLP:conf/icml/TanL19} and TinyNet~\cite{han2020model} employ so many resources in searching the optimal $(d^\star, w^\star, r^\star)$, while we do not have enough GPUs to reproduce their searching process. Thus, the results of EfficientNet and TinyNet are both drawn from \cite{han2020model}, and their search costs are estimated through the number of models they trained. For example, training an EfficientNet for $300$ epochs takes about $26$ hours with $8 \times $V100 GPUs, while TinyNet requires to train $60$ EfficientNet models from scratch. Hence, its search cost is about $520$ GPU days. Instead, our algorithm only spends about $\frac{1}{25}$ as much time as TinyNet on searching but achieves similar accuracy.

\begin{table}[t] \centering
	\caption{Pruning Results on ImageNet. The improvement on DenseNet-121 is marginal because of DenseNet-121's property. Detailed reasons are described in Section~\ref{exp:re_and_ana}.}
	\scalebox{0.76}{
		\begin{tabular}{llclccc}
			\toprule
			\multicolumn{1}{l|}{Algorithm} & \multicolumn{1}{c}{$\mathcal{D}$} & $\mathcal{W}$ & \multicolumn{1}{c|}{$\mathcal{R}$} & Accuracy & $Prr$ & $Frr$ \\
			\midrule
			\multicolumn{7}{c}{ResNet-50 (76.15\%)} \\
			\midrule
			\multicolumn{1}{l|}{$\mathcal{R}$-only~\cite{DBLP:journals/corr/HowardZCKWWAA17}} &  & \multicolumn{1}{l}{} & \multicolumn{1}{c|}{\checkmark} & 71.56\% & - & 0.50 \\
			\multicolumn{1}{l|}{$\mathcal{W}$-only~\cite{DBLP:conf/iccv/LiuLSHYZ17}} &  & \checkmark & \multicolumn{1}{l|}{} & 74.52\% & 0.50 & 0.50 \\
			\multicolumn{1}{l|}{DBP~\cite{DBLP:journals/corr/abs-1912-10178}} & \multicolumn{1}{c}{\checkmark} & \multicolumn{1}{l}{} & \multicolumn{1}{l|}{} & 73.92\% & 0.56 & 0.50 \\
			\multicolumn{1}{l|}{GAL$^\dagger$~\cite{DBLP:conf/cvpr/LinJYZCYHD19}} & \multicolumn{1}{c}{\checkmark} & \checkmark & \multicolumn{1}{l|}{} & 71.95\% & 0.17 & 0.43 \\
			\multicolumn{1}{l|}{FPGM$^\dagger$~\cite{he2019filter}} &  & \checkmark & \multicolumn{1}{l|}{} & 74.83\% & - & 0.54 \\
			\multicolumn{1}{l|}{PScratch$^\dagger$~\cite{DBLP:conf/aaai/WangZXZSZH20}} &  & \checkmark & \multicolumn{1}{l|}{} & 75.45\% & 0.64 & 0.50 \\
			\multicolumn{1}{l|}{HRank$^\dagger$~\cite{DBLP:conf/aaai/WangZXZSZH20}} &  & \checkmark & \multicolumn{1}{l|}{} & 74.98\% & 0.37 & 0.44 \\
			\multicolumn{1}{l|}{DHP~\cite{DBLP:conf/eccv/LiGZGT20}} &  & \checkmark & \multicolumn{1}{l|}{} & 75.45\% & 0.54 & 0.50 \\
			\multicolumn{1}{l|}{Ours} & \multicolumn{1}{c}{\checkmark} & \checkmark & \multicolumn{1}{c|}{\checkmark} & \textbf{75.90\%}& 0.53 & 0.50 \\
			\midrule
			\multicolumn{7}{c}{DenseNet-121 (75.01\%)} \\
			\midrule
			\multicolumn{1}{l|}{$\mathcal{R}$-only~\cite{DBLP:journals/corr/HowardZCKWWAA17}} &  & \multicolumn{1}{l}{} & \multicolumn{1}{c|}{\checkmark} & 73.07\% & - & 0.51 \\
			\multicolumn{1}{l|}{$\mathcal{W}$-only~\cite{DBLP:conf/iccv/LiuLSHYZ17}} &  & \checkmark & \multicolumn{1}{l|}{} & 73.58\% & 0.51 & 0.51 \\
			\multicolumn{1}{l|}{DBP~\cite{DBLP:journals/corr/abs-1912-10178}} & \multicolumn{1}{c}{\checkmark} & \multicolumn{1}{l}{} & \multicolumn{1}{l|}{} & 68.08\% & 0.66 & 0.37 \\
			\multicolumn{1}{l|}{Ours} & \multicolumn{1}{c}{\checkmark} & \checkmark & \multicolumn{1}{c|}{\checkmark} & \textbf{73.68\%} & 0.48 & 0.51 \\
			\bottomrule
		\end{tabular}
	}
%	\vspace{-0.07cm}
	\label{tab:imagenet}
\end{table}

\begin{table}[t] \centering
	\caption{Comparison with NAS-based model acceleration algorithms. They all take EfficientNet-B0 as the baseline model. GPU-days is measured with NVIDIA V100. Note that training an EfficientNet from scratch costs about 8.7 GPU days.}
	\scalebox{0.75}{
		\begin{tabular}{l|rrcr}
			\toprule
			\multirow{2}{*}{Algorithm} & \multirow{2}{*}{Params} & \multirow{2}{*}{FLOPs} & \multirow{2}{*}{Top-1/Top-5 Acc.} & Search Cost \\
			& & & & (GPU days) \\
			\midrule
			EfficientNet-B0 & 5.3M & 387M & 76.7\%/93.2\% & - \\
			TinyNet-A$^\dagger$ & 5.1M & 339M & \textbf{76.8\%}/\textbf{93.3\%} & $\sim$ 520   \\
			Ours & 5.1M & \textbf{314M} & \textbf{76.8\%}/\textbf{93.3\%} & 26  \\
			\midrule
			EfficientNet-B$^{-1}$ & 3.6M & 201M & 74.7\%/92.1\% & -   \\
			TinyNet-B$^\dagger$ & 3.7M & 202M & 75.0\%/92.2\% & $\sim$ 520   \\
			Ours & 3.6M & \textbf{198M} & \textbf{75.2\%}/\textbf{92.7\%} & 24  \\
			\midrule
			EfficientNet-B$^{-2}$ & 3.0M & \textbf{98M} & 70.5\%/89.5\% & -  \\
			TinyNet-C$^\dagger$ & 2.5M & 100M & 71.2\%/89.7\% & $\sim$ 520  \\
			Ours & 3.1M & \textbf{98M} & \textbf{71.6\%}/\textbf{89.9\%} & 21  \\
			\bottomrule
		\end{tabular}
	}
	\vspace{-0.07cm}
	\label{tab:nas}
\end{table}

\subsection{Ablation Study} \label{sec:ablation}
	
\begin{figure}  \centering
	\includegraphics[scale=0.55]{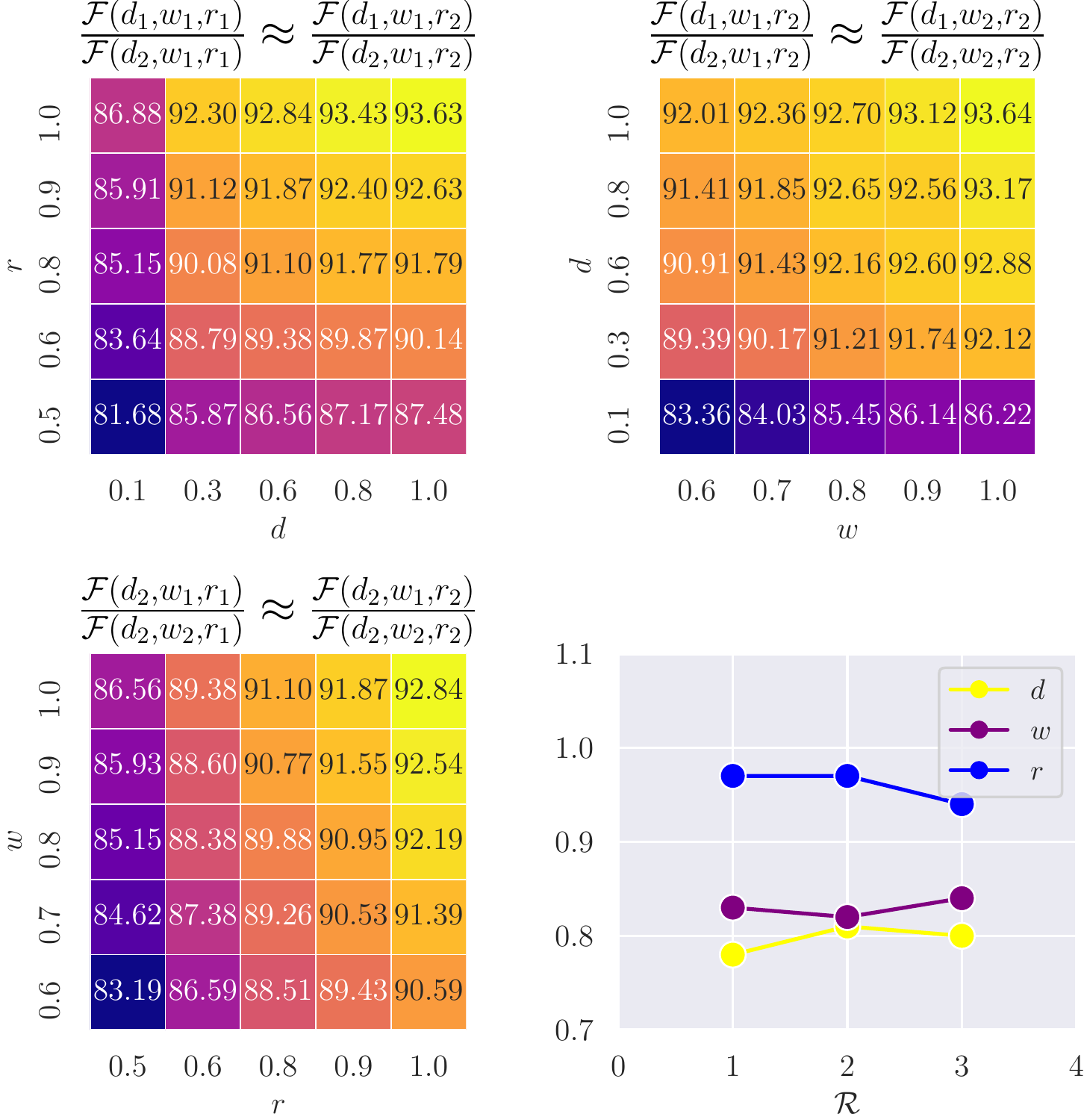}
	\caption{Accuracies of ResNets with different $(d, w, r)$ trained on CIFAR-10 (the first three sub-figures) and the predicted optimal $(d^\star, w^\star, r^\star)$ with different $\mathcal{R}$ in Eq.~\eqref{equ:exp1}.}
	\label{fig:sep}
%	\vspace{-0.3cm}
\end{figure}

\paragraph{Rank of $\Theta$:} In Section~\ref{sec:simplified-MAP}, our proposed MAP is
\begin{equation}
	\label{equ:exp1}
	\begin{aligned}
	\mathcal{F}(d, w, r; \Theta) = \sum_{q=1}^\mathcal{R} \mathcal{H}_{}(d; \vec{s_q})\mathcal{H}_{}(w; \vec{u_q})\mathcal{H}_{}(r; \vec{v_q}),
	\end{aligned}
\end{equation}
where the rank of $\Theta$ is less than or equal to $\mathcal{R}$. Experimentally, we find that $\mathcal{R} = 1$ works well in most cases. To further explore this interesting phenomenon, ResNets with different $(d, w, r)$ are trained on CIFAR-10. The base model (i.e., $(d, w, r) = (1.0, 1.0, 1.0)$) is ResNet-32 with images of size $32$, and the results are plotted in Figure~\ref{fig:sep}. Observations from the first three sub-figures are shown in their titles. We can deduce from these observations\footnote{\fontsize{8.4pt}{\baselineskip}\selectfont More results and the proof are put in our supplementary material.}:
\begin{equation}
	\label{equ:exp2}
	\begin{aligned}
	\mathcal{F}(d, w, r; \Theta) \approx \mathcal{H}_{}(d; \vec{s_q})\mathcal{H}_{}(w; \vec{u_q})\mathcal{H}_{}(r; \vec{v_q}),
	\end{aligned}
\end{equation}
which coincides with Eq.~\eqref{equ:exp1} once $\mathcal{R} = 1$. In other words, three variables $(d, w, r)$ in the MAP can be approximately separated from each other. 
We also test the MAP with different $\mathcal{R}$, as shown in the 4\textit{th} sub-figure of Figure~\ref{fig:sep}. The MAP with larger $\mathcal{R}$ yields similar $(d^\star, w^\star, r^\star)$ to that of $\mathcal{R}=1$, which also indicates that $\mathcal{R}=1$ is enough for obtaining a well-performed MAP.

\paragraph{Other Methods of Avoiding Overfitting:}
	
Besides restricting the rank of $\Theta$, we also try two extra methods of avoiding overfitting, i.e., decreasing the degree $\mathcal{K}$ of polynomials and applying regression with regularization terms. The results are reported in Table~\ref{tab:different-MAP1}.  Specifically, a set of $13$ items $(d_n, w_n, r_n, a_n)$ is used to fit the MAP, and a set of the other $80$ items is used for evaluation. All data is collected with ResNet-56 trained on CIFAR-10. Training error and evaluation error are both reported. As we can see, normal polynomial regression induces severe overfitting and high evaluation loss, and $\ell_2$-regularization has a limited effect on dealing with the overfitting issue. Still, lowering the degree of polynomials is not a wise choice because it makes the polynomials fail to converge even on training data.
Instead, our specialized MAP shows a lower error rate on both training data ($0.08\%$) and evaluation data ($0.25\%$).
	%\vspace{-0.2cm}

\paragraph{Influence of Polynomials' Degree:}

\begin{figure}  \centering
	\includegraphics[scale=0.90]{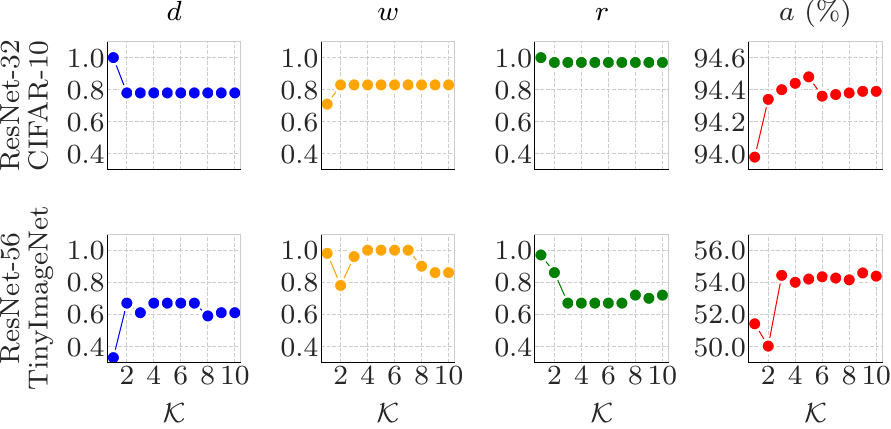}
	\caption{The predicted optimal $(d^\star, w^\star, r^\star)$ and corresponding pruning results with different $\mathcal{K}$. For all $\mathcal{K} \ge 3$, the MAP's predicted optimal $(d^\star, w^\star, r^\star)$ are very similar. Users do not need to bother to choose $\mathcal{K}$ carefully.}
	\label{fig:degree}
	%		\vspace{-0.2cm}
\end{figure}

Figure~\ref{fig:degree} shows the pruning results when adjusting the MAP's degree $\mathcal{K}$. Especially, the polynomial regression degrades to linear regression when $\mathcal{K}=1$. It turns out that for polynomials with $\mathcal{K} \le 2$, the predicted optimal $(d^\star, w^\star, r^\star)$ actually leads to a sub-optimal pruning policy, which indicates that the MAP is too simple to use. For polynomials with $\mathcal{K} \ge 3$, all MAPs generate similar predictions about the optimal $(d, w, r)$, i.e., $(0.78, 0.82, 0.98)$ for ResNet-32 trained on CIFAR-10 and $(0.65, 1.0, 0.63)$ for ResNet-56 trained on TinyImageNet. These results corroborate that our algorithm is relatively robust with respect to different degrees $\mathcal{K}$, so practitioners do not need to choose the polynomial degree carefully.
	
\begin{table}[]
	\centering
	\caption{MAP's regression results with lower degree or regularization. $\mathcal{K}$ means the highest degree for each variable in $(d, w, r)$. $\ell_2$ is the coefficient of the regularization term, and 0 indicates no regularization.}
	\label{tab:different-MAP1}
	\scalebox{0.86}{
		\setlength{\tabcolsep}{4.mm}{
			\begin{tabular}{c|cccc}
				\toprule
				\multicolumn{1}{c|}{Type} & $\mathcal{K}$ & $\ell_2$ & Train Err.& Eval Err. \\
				\midrule
				\multirow{6}{*}{Normal} & 1 & 0 & 2.66\% & 2.97\% \\
				\multirow{6}{*}{Polynomial}& 2 & 0 & 1.62\% & 2.25\% \\
				& 3 & 0 & 0.28\% & 1.28\% \\
				& 5 & 0 & 0.02\% & 2.31\% \\
				& 5 & $10^{-3}$ & 0.02\% & 2.28\% \\
				& 10 & 0 & 0.01\% & 2.58\% \\
				& 10 & $10^{-3}$  & 0.02\% & 2.32\% \\
				\midrule
				\multirow{2}{*}{Ours} & 3 & 0 & 0.14\% & 0.33\% \\
				& 5 & 0 & 0.08\% & \textbf{0.25\%} \\
				\bottomrule
		\end{tabular}}
	}
\end{table}

\begin{figure}  \centering
	\includegraphics[scale=1.0]{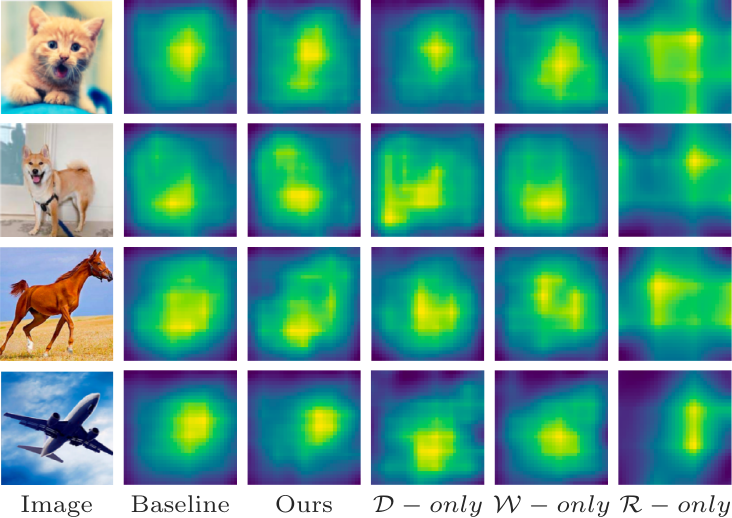}
	\caption{Visualization of last layer's feature maps from different models. The baseline model is a pre-trained ResNet-56. Four different pruning policies are tested, and our pruned model's feature maps look most like those of the baseline model.}
	\label{fig:visualization}
			\vspace{-0.35cm}
\end{figure}
	
%	\vspace{-0.2cm}
\subsection{Case Study}

%	\vspace{-0.1cm}
\paragraph{Visualization of Feature Maps for Different Pruning Policies:}

In order to further understand why pruning the three dimensions simultaneously yields better results than pruning only one, Figure~\ref{fig:visualization} compares the feature maps for models with different pruning policies. Specifically, all models are pruned from the same baseline model --- ResNet-56 pre-trained on CIFAR-10. Input images are randomly chosen from the Internet. For visualization, we extract the last convolutional layer's feature maps and compute their mean absolute values across channels. 
Figure~\ref{fig:visualization} shows that the feature maps our pruned model outputs look most similar to those of the original model. 
This finding reveals that our algorithm preserves most information of the original model by pruning the three dimensions comprehensively.

\section{Conclusion}

In this paper, we proposed a novel pruning framework which prunes a pre-trained model along three dimensions, i.e., depth, width, and resolution, comprehensively. 
Remarkably, our framework can determine the optimal values for these three dimensions through modeling the relationships between 
the model's accuracy and depth/width/resolution into a polynomial regression and subsequently solving an optimization problem.
The extensive experimental results demonstrate that the proposed pruning algorithm outperforms state-of-the-art pruning algorithms under  
a comparable computational budget. In contrast with NAS-based methods, we generated the pruned models that are superior to the NAS-searched models 
with a much reduced computational cost.

\section*{Acknowledgements}
	
This work was supported in part by The National Key Research and Development Program of China (Grant No. 2018AAA0101400), in part by The National Nature Science Foundation of China (Grant Nos: 62036009, U1909203, 61936006), and in part by Innovation Capability Support Program of Shaanxi (Program No. 2021TD-05).

\bibliographystyle{icml2021}
\bibliography{MAP}

	\appendix

\begin{table*}[]
	\caption{Accuracies (\%) of ResNets and DenseNets on CIFAR-10 with different depths ($d$), widths ($w$) and resolutions ($r$). The base models are ResNet-32 and DenseNet-40.}
	\begin{center}
		\begin{subtable}{0.32\textwidth}
			\caption{$\frac{\mathcal{F}(d_2, w_1, r_1)}{\mathcal{F}(d_2, w_2, r_1)} \approx \frac{\mathcal{F}(d_2, w_1, r_2)}{\mathcal{F}(d_2, w_2, r_2)}$}
			\scalebox{0.72}{
				\begin{tabular}{c|ccccc}
					\toprule
					\multicolumn{6}{c}{ResNet ($d=1.0$)} \\
					\midrule
					\diagbox{$w$}{$r$} & 1.00 & 0.87 & 0.75 & 0.62 & 0.50 \\
					\midrule
					%							0.50 & 89.25 & 88.25 & 87.46 & 85.35 & 82.24 \\
					0.60 & 90.59 & 89.43 & 88.51 & 86.59 & 83.19 \\
					0.70 & 91.39 & 90.53 & 89.26 & 87.38 & 84.62 \\
					0.80 & 92.19 & 90.95 & 89.88 & 88.38 & 85.15 \\
					0.90 & 92.54 & 91.55 & 90.77 & 88.60 & 85.93 \\
					1.00 &  92.84 & 91.87 & 91.10 & 89.38 & 86.56 \\
					\midrule
					\midrule
					\multicolumn{6}{c}{DenseNet ($d=1.0$)} \\
					\midrule
					\diagbox{$w$}{$r$} & 1.00 & 0.87 & 0.75 & 0.62 & 0.50 \\
					\midrule
					%							0.500 & 89.85 & 89.41 & 88.70 & 87.11 & 85.65 \\
					0.60 & 90.82 & 90.48 & 89.62 & 88.04 & 85.63 \\
					0.70 & 91.54 & 90.98 & 90.19 & 88.79 & 86.75 \\
					0.80 & 91.99 & 91.57 & 90.59 & 90.01 & 87.67 \\
					0.90 & 92.74 & 92.08 & 91.19 & 90.56 & 88.12 \\
					1.00 & 93.09 & 92.25 & 92.05 & 90.68 & 88.38 \\
					\bottomrule
			\end{tabular}}
		\end{subtable}
		\begin{subtable}{0.32\textwidth}
			\caption{$\frac{\mathcal{F}(d_1, w_1, r_1)}{\mathcal{F}(d_2, w_1, r_1)} \approx \frac{\mathcal{F}(d_1, w_1, r_2)}{\mathcal{F}(d_2, w_1, r_2)}$}
			\scalebox{0.72}{
				\begin{tabular}{c|ccccc}
					\toprule
					\multicolumn{6}{c}{ResNet ($w = 1.0$)} \\
					\midrule
					\diagbox{$d$}{$r$} & 1.00 & 0.87 & 0.75 & 0.62 & 0.50 \\
					\midrule
					0.11  & 86.88 & 85.91 & 85.15 & 83.64 & 81.68 \\
					0.33 & 92.30 & 91.12 & 90.08 & 88.79 & 85.87 \\
					0.55 & 92.84 & 91.87 & 91.10 & 89.38 & 86.56 \\
					0.77 & 93.43 & 92.40 & 91.77 & 89.87 & 87.17 \\
					1.00 & 93.63 & 92.63 & 91.79 & 90.14 & 87.48 \\
					\midrule
					\midrule
					\multicolumn{6}{c}{DenseNet ($w = 1.0$)} \\
					\midrule
					\diagbox{$d$}{$r$} & 1.00 & 0.87 & 0.75 & 0.62 & 0.50 \\
					\midrule
					0.20 & 88.16 & 87.64 & 86.77 & 85.50 & 83.83 \\
					0.40 & 92.00 & 91.22 & 90.32 & 89.15 & 86.89 \\
					0.60 & 93.03 & 92.06 & 91.85 & 90.63 & 88.42 \\
					0.80 & 93.80 & 93.21 & 92.78 & 91.74 & 89.25 \\
					1.00 & 94.53 & 93.75 & 93.69 & 92.21 & 89.84 \\
					\bottomrule
			\end{tabular}}
		\end{subtable}
		\begin{subtable}{0.32\textwidth}
			\caption{$\frac{\mathcal{F}(d_1, w_1, r_2)}{\mathcal{F}(d_2, w_1, r_2)} \approx \frac{\mathcal{F}(d_1, w_2, r_2)}{\mathcal{F}(d_2, w_2, r_2)} $}
			\scalebox{0.72}{
				\begin{tabular}{c|ccccc}
					\toprule
					\multicolumn{6}{c}{ResNet ($r = 1.0$)} \\
					\midrule
					\diagbox{$w$}{$d$} & 0.11 & 0.33 & 0.55 & 0.77 & 1.00 \\
					\midrule
					%										0.500  & 82.50 &  88.51 & 89.89 & 90.51 & 90.91 \\
					0.60  & 83.36 & 89.39 & 90.91 & 91.41 & 92.01 \\
					0.70  & 84.03 & 90.17 & 91.43 & 91.85 & 92.36 \\
					0.80  & 85.45 & 91.21 & 92.16 & 92.65 & 92.7 \\
					0.90  & 86.14 & 91.74 & 92.60 &  92.56 & 93.12 \\
					1.00 &  86.22 & 92.12 & 92.88 & 93.17 & 93.64 \\
					\midrule
					\midrule
					\multicolumn{6}{c}{DenseNet ($r = 1.0$)} \\
					\midrule
					\diagbox{$w$}{$d$} & 0.20 & 0.40 & 0.60 & 0.80 & 1.00 \\
					\midrule
					%										0.500 & 83.53 & 87.85 & 89.10 & 91.00 & 91.91 \\
					0.60 & 84.51 & 88.82 & 89.95 & 91.40 & 92.70 \\
					0.70 & 85.43 & 89.26 & 90.51 & 92.35 & 92.83 \\
					0.80 & 86.92 & 90.75 & 91.07 & 93.01 & 93.66 \\
					0.90 & 87.88 & 91.74 & 91.46 & 93.14 & 93.86 \\
					1.00 & 88.16 & 92.00 & 93.03 & 93.80 & 94.53 \\
					\bottomrule
			\end{tabular}}
		\end{subtable}
	\end{center}
	\label{tab:supp1}
\end{table*}

\section{Why the Rank of $\Theta$ Is 1}
\subsection{Experiments}

In our original paper, we propose a model accuracy predictor (MAP):
\begin{equation}
	\label{equ:supp_exp1}
	\begin{aligned}
		\mathcal{F}(d, w, r; \Theta) = \sum_{q=1}^\mathcal{R} \mathcal{H}_{}(d; \vec{s_q})\mathcal{H}_{}(w; \vec{u_q})\mathcal{H}_{}(r; \vec{v_q}),
	\end{aligned}
\end{equation}
where $\mathcal{H}$ represents a univariate polynomial, and $\mathcal{R}$ means the rank of $\Theta$. Pruning experiments shown in our original paper are done with $\mathcal{R}=1$ because we find that $\mathcal{R} = 1$ is enough for an accurate approximation of MAP. In this case, Eq.~\eqref{equ:supp_exp1} becomes:
\begin{equation}
	\label{equ:supp_exp2}
	\begin{aligned}
		\mathcal{F}(d, w, r; \Theta) = \mathcal{H}_{}(d; \vec{s})\mathcal{H}_{}(w; \vec{u})\mathcal{H}_{}(r; \vec{v}).
	\end{aligned}
\end{equation}
We assume that $\mathcal{R} = 1$ works well because it accords with the real distribution of $\mathcal{F}(d, w, r)$. To further verify the assumption. we design some experiments to show the relations between the model's accuracy and $(d, w, r)$. Specifically, we train many ResNets and DenseNets with different $(d, w, r)$, and the results are in Table~\ref{tab:supp1} --- part of them have also been reported in the original paper. The observations are shown in their subtitles, from which we can draw the same conclusion as Eq.~\eqref{equ:supp_exp2}.

\subsection{Proof}
Omitting all subscripts of $d_1$, $w_1$, and $r_1$ of the equations in Table~\ref{tab:supp1}, we have:
\begin{equation}
	\label{equ:supp2-1}
	\begin{aligned}
		\frac{\mathcal{F}(d_2, w, r)}{\mathcal{F}(d_2, w_2, r)} &\approx \frac{\mathcal{F}(d_2, w, r_2)}{\mathcal{F}(d_2, w_2, r_2)} \\
		\Rightarrow  \mathcal{F}(d_2, w, r) &\approx \frac{\mathcal{F}(d_2, w_2, r)\mathcal{F}(d_2, w, r_2)}{\mathcal{F}(d_2, w_2, r_2)}, \\
	\end{aligned}
\end{equation}
\begin{equation}
	\label{equ:supp2-2}
	\begin{aligned}
		\frac{\mathcal{F}(d, w, r)}{\mathcal{F}(d_2, w, r)} &\approx \frac{\mathcal{F}(d, w, r_2)}{\mathcal{F}(d_2, w, r_2)}  \\
		\Rightarrow \mathcal{F}(d, w, r) &\approx \frac{\mathcal{F}(d_2, w, r)\mathcal{F}(d, w, r_2)}{\mathcal{F}(d_2, w, r_2)}, \\
	\end{aligned}
\end{equation}

\begin{equation}
	\label{equ:supp2-3}
	\begin{aligned}
		\qquad \frac{\mathcal{F}(d, w, r_2 )}{\mathcal{F}(d_2, w, r_2)} &= \frac{\mathcal{F}(d, w_2, r_2)}{\mathcal{F}(d_2, w_2, r_2)}  \\
		\Rightarrow \mathcal{F}(d, w, r_2 ) &= \frac{\mathcal{F}(d_2, w, r_2)\mathcal{F}(d, w_2, r_2)}{\mathcal{F}(d_2, w_2, r_2)}.\\
	\end{aligned}
\end{equation}
Substituting Eq.~\eqref{equ:supp2-1} and Eq.~\eqref{equ:supp2-3} into Eq.~\eqref{equ:supp2-2}, we have:
\begin{equation}
	\label{equ:supp3-1}
	\begin{aligned}
		\mathcal{F}(d, w, r) &\approx \frac{\mathcal{F}(d, w_2, r_2)\mathcal{F}(d_2, w, r_2)\mathcal{F}(d_2, w_2, r)}{\mathcal{F}(d_2, w_2, r_2)^2} \\ 
		&= \frac{\mathcal{H}(d;\vec{s})  \mathcal{H}(w;\vec{u})  \mathcal{H}(r;\vec{v})}{{\mathcal{F}(d_2, w_2, r_2)^2}},
	\end{aligned}
\end{equation}
where $\frac{1}{\mathcal{F}(d_2, w_2, r_2)^2}$ is a constant and can be merged into any $\mathcal{H}$. Thus, Eq.~\eqref{equ:supp3-1} can be re-formulated as:
\begin{equation}
	\label{equ:supp3-2}
	\begin{aligned}
		\mathcal{F}(d, w, r) \approx \mathcal{H}(d;\vec{s})  \mathcal{H}(w;\vec{u})  \mathcal{H}(r;\vec{v}),
	\end{aligned}
\end{equation}
which complies with Eq.~\eqref{equ:supp_exp2}.

\end{document}

\twocolumn[
	\icmltitle{The Supplementary Material for ``Accelerate CNNs from Three Dimensions: A Comprehensive Pruning Framework''}
	
	\vskip 0.1in
	
	]

	\appendix
	
	\begin{table*}[]
		\caption{Accuracies (\%) of ResNets and DenseNets on CIFAR-10 with different depths ($d$), widths ($w$) and resolutions ($r$). The base models are ResNet-32 and DenseNet-40.}
		\begin{center}
			\begin{subtable}{0.32\textwidth}
				\caption{$\frac{\mathcal{F}(d_2, w_1, r_1)}{\mathcal{F}(d_2, w_2, r_1)} \approx \frac{\mathcal{F}(d_2, w_1, r_2)}{\mathcal{F}(d_2, w_2, r_2)}$}
				\scalebox{0.72}{
					\begin{tabular}{c|ccccc}
						\toprule
						\multicolumn{6}{c}{ResNet ($d=1.0$)} \\
						\midrule
						\diagbox{$w$}{$r$} & 1.00 & 0.87 & 0.75 & 0.62 & 0.50 \\
						\midrule
						%							0.50 & 89.25 & 88.25 & 87.46 & 85.35 & 82.24 \\
						0.60 & 90.59 & 89.43 & 88.51 & 86.59 & 83.19 \\
						0.70 & 91.39 & 90.53 & 89.26 & 87.38 & 84.62 \\
						0.80 & 92.19 & 90.95 & 89.88 & 88.38 & 85.15 \\
						0.90 & 92.54 & 91.55 & 90.77 & 88.60 & 85.93 \\
						1.00 &  92.84 & 91.87 & 91.10 & 89.38 & 86.56 \\
						\midrule
						\midrule
						\multicolumn{6}{c}{DenseNet ($d=1.0$)} \\
						\midrule
						\diagbox{$w$}{$r$} & 1.00 & 0.87 & 0.75 & 0.62 & 0.50 \\
						\midrule
						%							0.500 & 89.85 & 89.41 & 88.70 & 87.11 & 85.65 \\
						0.60 & 90.82 & 90.48 & 89.62 & 88.04 & 85.63 \\
						0.70 & 91.54 & 90.98 & 90.19 & 88.79 & 86.75 \\
						0.80 & 91.99 & 91.57 & 90.59 & 90.01 & 87.67 \\
						0.90 & 92.74 & 92.08 & 91.19 & 90.56 & 88.12 \\
						1.00 & 93.09 & 92.25 & 92.05 & 90.68 & 88.38 \\
						\bottomrule
				\end{tabular}}
			\end{subtable}
			\begin{subtable}{0.32\textwidth}
				\caption{$\frac{\mathcal{F}(d_1, w_1, r_1)}{\mathcal{F}(d_2, w_1, r_1)} \approx \frac{\mathcal{F}(d_1, w_1, r_2)}{\mathcal{F}(d_2, w_1, r_2)}$}
				\scalebox{0.72}{
					\begin{tabular}{c|ccccc}
						\toprule
						\multicolumn{6}{c}{ResNet ($w = 1.0$)} \\
						\midrule
						\diagbox{$d$}{$r$} & 1.00 & 0.87 & 0.75 & 0.62 & 0.50 \\
						\midrule
						0.11  & 86.88 & 85.91 & 85.15 & 83.64 & 81.68 \\
						0.33 & 92.30 & 91.12 & 90.08 & 88.79 & 85.87 \\
						0.55 & 92.84 & 91.87 & 91.10 & 89.38 & 86.56 \\
						0.77 & 93.43 & 92.40 & 91.77 & 89.87 & 87.17 \\
						1.00 & 93.63 & 92.63 & 91.79 & 90.14 & 87.48 \\
						\midrule
						\midrule
						\multicolumn{6}{c}{DenseNet ($w = 1.0$)} \\
						\midrule
						\diagbox{$d$}{$r$} & 1.00 & 0.87 & 0.75 & 0.62 & 0.50 \\
						\midrule
						0.20 & 88.16 & 87.64 & 86.77 & 85.50 & 83.83 \\
						0.40 & 92.00 & 91.22 & 90.32 & 89.15 & 86.89 \\
						0.60 & 93.03 & 92.06 & 91.85 & 90.63 & 88.42 \\
						0.80 & 93.80 & 93.21 & 92.78 & 91.74 & 89.25 \\
						1.00 & 94.53 & 93.75 & 93.69 & 92.21 & 89.84 \\
						\bottomrule
				\end{tabular}}
			\end{subtable}
			\begin{subtable}{0.32\textwidth}
				\caption{$\frac{\mathcal{F}(d_1, w_1, r_2)}{\mathcal{F}(d_2, w_1, r_2)} \approx \frac{\mathcal{F}(d_1, w_2, r_2)}{\mathcal{F}(d_2, w_2, r_2)} $}
				\scalebox{0.72}{
					\begin{tabular}{c|ccccc}
						\toprule
						\multicolumn{6}{c}{ResNet ($r = 1.0$)} \\
						\midrule
						\diagbox{$w$}{$d$} & 0.11 & 0.33 & 0.55 & 0.77 & 1.00 \\
						\midrule
						%										0.500  & 82.50 &  88.51 & 89.89 & 90.51 & 90.91 \\
						0.60  & 83.36 & 89.39 & 90.91 & 91.41 & 92.01 \\
						0.70  & 84.03 & 90.17 & 91.43 & 91.85 & 92.36 \\
						0.80  & 85.45 & 91.21 & 92.16 & 92.65 & 92.7 \\
						0.90  & 86.14 & 91.74 & 92.60 &  92.56 & 93.12 \\
						1.00 &  86.22 & 92.12 & 92.88 & 93.17 & 93.64 \\
						\midrule
						\midrule
						\multicolumn{6}{c}{DenseNet ($r = 1.0$)} \\
						\midrule
						\diagbox{$w$}{$d$} & 0.20 & 0.40 & 0.60 & 0.80 & 1.00 \\
						\midrule
						%										0.500 & 83.53 & 87.85 & 89.10 & 91.00 & 91.91 \\
						0.60 & 84.51 & 88.82 & 89.95 & 91.40 & 92.70 \\
						0.70 & 85.43 & 89.26 & 90.51 & 92.35 & 92.83 \\
						0.80 & 86.92 & 90.75 & 91.07 & 93.01 & 93.66 \\
						0.90 & 87.88 & 91.74 & 91.46 & 93.14 & 93.86 \\
						1.00 & 88.16 & 92.00 & 93.03 & 93.80 & 94.53 \\
						\bottomrule
				\end{tabular}}
			\end{subtable}
		\end{center}
		\label{tab:supp1}
	\end{table*}
	
	\section{Why the Rank of $\Theta$ Is 1}
	\subsection{Experiments}
	
	In our original paper, we propose a model accuracy predictor (MAP):
	\begin{equation}
	\begin{aligned}
	\mathcal{F}(d, w, r; \Theta) = \sum_{q=1}^\mathcal{R} \mathcal{H}_{}(d; \vec{s_q})\mathcal{H}_{}(w; \vec{u_q})\mathcal{H}_{}(r; \vec{v_q}),
	\end{aligned}
	\end{equation}
	where $\mathcal{H}$ represents a univariate polynomial, and $\mathcal{R}$ means the rank of $\Theta$. Pruning experiments shown in our original paper are done with $\mathcal{R}=1$ because we find that $\mathcal{R} = 1$ is enough for an accurate approximation of MAP. In this case, Eq.~\eqref{equ:exp1} becomes:
	\begin{equation}
	\label{equ:exp2}
	\begin{aligned}
	\mathcal{F}(d, w, r; \Theta) = \mathcal{H}_{}(d; \vec{s})\mathcal{H}_{}(w; \vec{u})\mathcal{H}_{}(r; \vec{v}).
	\end{aligned}
	\end{equation}
	We assume that $\mathcal{R} = 1$ works well because it accords with the real distribution of $\mathcal{F}(d, w, r)$. To further verify the assumption. we design some experiments to show the relations between the model's accuracy and $(d, w, r)$. Specifically, we train many ResNets and DenseNets with different $(d, w, r)$, and the results are in Table~\ref{tab:supp1} --- part of them have also been reported in the original paper. The observations are shown in their subtitles, from which we can draw the same conclusion as Eq.~\eqref{equ:exp2}.
	
	\subsection{Proof}
	Omitting all subscripts of $d_1$, $w_1$, and $r_1$ of the equations in Table~\ref{tab:supp1}, we have:
	\begin{equation}
	\label{equ:supp2-1}
	\begin{aligned}
	\frac{\mathcal{F}(d_2, w, r)}{\mathcal{F}(d_2, w_2, r)} &\approx \frac{\mathcal{F}(d_2, w, r_2)}{\mathcal{F}(d_2, w_2, r_2)} \\
	\Rightarrow  \mathcal{F}(d_2, w, r) &\approx \frac{\mathcal{F}(d_2, w_2, r)\mathcal{F}(d_2, w, r_2)}{\mathcal{F}(d_2, w_2, r_2)}, \\
	\end{aligned}
	\end{equation}
	\begin{equation}
	\label{equ:supp2-2}
	\begin{aligned}
	\frac{\mathcal{F}(d, w, r)}{\mathcal{F}(d_2, w, r)} &\approx \frac{\mathcal{F}(d, w, r_2)}{\mathcal{F}(d_2, w, r_2)}  \\
	\Rightarrow \mathcal{F}(d, w, r) &\approx \frac{\mathcal{F}(d_2, w, r)\mathcal{F}(d, w, r_2)}{\mathcal{F}(d_2, w, r_2)}, \\
	\end{aligned}
	\end{equation}
	
	\begin{equation}
	\label{equ:supp2-3}
	\begin{aligned}
	\qquad \frac{\mathcal{F}(d, w, r_2 )}{\mathcal{F}(d_2, w, r_2)} &= \frac{\mathcal{F}(d, w_2, r_2)}{\mathcal{F}(d_2, w_2, r_2)}  \\
	\Rightarrow \mathcal{F}(d, w, r_2 ) &= \frac{\mathcal{F}(d_2, w, r_2)\mathcal{F}(d, w_2, r_2)}{\mathcal{F}(d_2, w_2, r_2)}.\\
	\end{aligned}
	\end{equation}
	Substituting Eq.~\eqref{equ:supp2-1} and Eq.~\eqref{equ:supp2-3} into Eq.~\eqref{equ:supp2-2}, we have:
	\begin{equation}
	\label{equ:supp3-1}
	\begin{aligned}
	\mathcal{F}(d, w, r) &\approx \frac{\mathcal{F}(d, w_2, r_2)\mathcal{F}(d_2, w, r_2)\mathcal{F}(d_2, w_2, r)}{\mathcal{F}(d_2, w_2, r_2)^2} \\ 
	&= \frac{\mathcal{H}(d;\vec{s})  \mathcal{H}(w;\vec{u})  \mathcal{H}(r;\vec{v})}{{\mathcal{F}(d_2, w_2, r_2)^2}},
	\end{aligned}
	\end{equation}
	where $\frac{1}{\mathcal{F}(d_2, w_2, r_2)^2}$ is a constant and can be merged into any $\mathcal{H}$. Thus, Eq.~\eqref{equ:supp3-1} can be re-formulated as:
	\begin{equation}
	\label{equ:supp3-2}
	\begin{aligned}
	\mathcal{F}(d, w, r) \approx \mathcal{H}(d;\vec{s})  \mathcal{H}(w;\vec{u})  \mathcal{H}(r;\vec{v}),
	\end{aligned}
	\end{equation}
	which complies with Eq.~\eqref{equ:exp2}.

	\bibliographystyle{icml2021}
	%\bibliography{MAP}